# Distantly Labeling Data for Large Scale Cross-Document Coreference

Sameer Singh, Michael Wick, and Andrew McCallum

Department of Computer Science
University of Massachusetts
Amherst MA 01003
{sameer,mwick,mccallum}@cs.umass.edu

**Abstract.** Cross-document coreference, the problem of resolving entity mentions across multi-document collections, is crucial to automated knowledge base construction and data mining tasks. However, the scarcity of large labeled data sets has hindered supervised machine learning research for this task. In this paper we develop and demonstrate an approach based on "distantly-labeling" a data set from which we can train a discriminative cross-document coreference model. In particular we build a dataset of more than a million people mentions extracted from 3.5 years of New York Times articles, leverage Wikipedia for distant labeling with a generative model (and measure the reliability of such labeling); then we train and evaluate a conditional random field coreference model that has factors on cross-document entities as well as mention-pairs. This coreference model obtains high accuracy in resolving mentions and entities that are not present in the training data, indicating applicability to non-Wikipedia data. Given the large amount of data, our work is also an exercise demonstrating the scalability of our approach.

## 1 Introduction

Given a collection of mentions of people or other entities extracted from a body of text, *coreference* or *entity resolution* consists of partitioning (or clustering) the mentions such that all mentions within a partition refer to the same underlying entity. Coreference is vital for many down-stream semantic analysis and knowledge discovery tasks [6]. Yet despite being extensively studied, considerable challenges remain. In particular, while significant progress has been made in *within*-document coreference [22,4,9,15] (resolving mentions from inside a single document), the larger problem of *cross*-document coreference (resolving mentions from across a collection of many documents) has received significantly less attention. We hypothesize two reasons for this discrepancy. First, there is a scarcity of substantial datasets labeled for cross-document coreference. Second, both the data and the hypothesis space are as a consequence larger and more difficult to manage.

One approach to coping with the lack of training data is to employ unsupervised methods, where weights, thresholds, or priors are set by hand. Some existing methods combine a clustering procedure with thresholded distance function over entities [1,14]. More recently, generative and non-parametric Bayesian clustering techniques have been proposed as a way to circumvent the need for labeled data [15]. Unfortunately, these unsupervised methods tend to perform considerably worse than supervised methods, making labeled data essential for achieving state-of-the art performance [4].

Of course another alternative is to manually label a corpus. However, this presents more difficulties for cross-document coreference than for within-document coreference. For cross-document coreference, the number of mentions and entities is typically large, and thus the space of partitions immense—making the labeling process cumbersome, with high uncertainty about the true number of entities. Faced with these difficulties cross-document labelers often use tools that allow them to query and label *pairs* of mentions at a time; however, this process scales poorly, has high cognitive load, and results in transitivity violations that must be resolved in a subsequent step. Furthermore, information from the immediate context of the mention is often not sufficient to resolve them, requiring additional search or investigation.

We present an alternative approach that supports supervised learning, while avoiding the need for human labeling effort. In addition to the unlabeled corpus of mentions, we leverage readily-available supplementary data—data which does not directly provide the required labels, but which is distantly related to those labels needed for the coreference task at hand. We process the original data together with this *distantly-labeled* data [31] to automatically label the original relevant corpus. This is a type of *indirect supervision* by alignment [3]. In this paper we develop and demonstrate distant-labeling for coreference of over a million mentions from 3.5 years of New York Times newspaper articles by using Wikipedia as distantly-labeled data. We present a generative probabilistic model that performs the alignment with 92% accuracy.

We also present a sophisticated conditional random field model of coreference that includes factors over entities, as well as traditional mention-mention pairs. The model is trained on the distantly-labeled data, and evaluated on unseen mentions and entities. We address the challenge of scaling up this model to our massive data set by using a family of Metropolis-Hastings proposal distributions that use canopies [20] to efficiently explore the hypothesis space. Our experiments show that both learning and inference can be performed in less than 10 hours on a single CPU, even though the coreference hypothesis space is exponential in the millions of cross-document mentions present in the corpus.

The rest of the paper is organized as follows: Section 2 defines and motivates the problem of cross-document coreference. In Section 3 we describe and evaluate our proposed approach for generating training data using Wikipedia. The cross-document coreference model trained on this data is described in Section 4. In Section 5 we explore related work. We conclude and lay out a number of ideas for future work in Section 6.

## 2   Problem Definition

The problem of cross-document coreference is to identify the sets of mention strings that refer to the same underlying entity, where the number of entities is not known. The source of the mentions may be a single document, in which case the task is within-document coreference. The number of mentions (and underlying entities) in each document is usually in the hundreds. The difficulty of the task arises from large hypothesis space (exponential in the number of mentions) and from challenges in resolving nominal and pronominal mentions to the correct named mentions. Usually, named mentions are not ambiguous in within-document coreference.

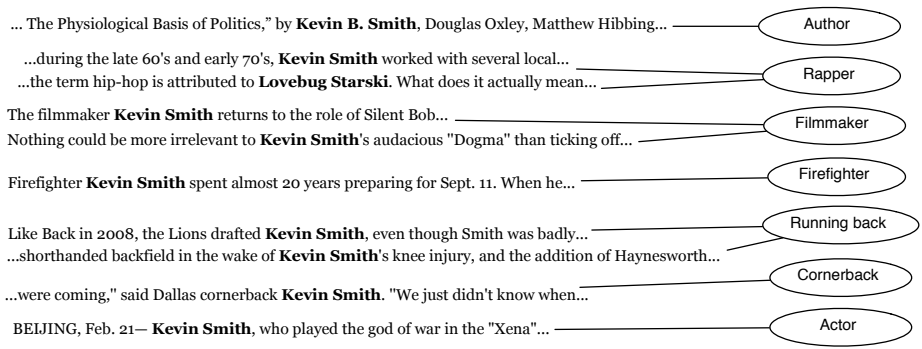

**Fig. 1.** Example of a set of ambiguous mention strings from news articles and their true entities.

The problem we study in this paper is that of *cross-document* coreference, where the source of mentions is a large collection of documents. The same sources of ambiguity as *within-doc* coreference exist in this scenario also. Furthermore, the number of mentions and entities are typically much larger, and for some corpora it can be in the millions, making the hypothesis space of all possible clustering exponentially larger than that of *within-doc*. Additionally, often there exists ambiguity even in the named entities since the same string can refer to multiple entities in different document, and multiple distinct strings may refer to the same entity in different documents.

We show an example of some ambiguities in Fig 1. The most common problem in cross-document coreference is resolving people that have the exact *same name*. The example above contains an example of such entities that are in the same general category (football), making the problem more difficult. Another common ambiguity is that of *alternate* names, where the same entity is referred to variations on the same name (such as "Richard/Dick"). The figure also shows an example of *renaming* ambiguity ("Lovebug Starski" may be mentioned as "Kevin Smith"), which is an extreme case of alternate names. Rare *singleton* entities (like the firefighter), who may appear only once in the whole corpus, are also often difficult to isolate. Our approach aims to address all these various challenges.

## 3 Distant-Labels for Coreference

Wikipedia[1] pages contain historical and biographical descriptions of a large number of entities, which we use as the external source of "distant labels". These pages are used to identify the entities that mention strings from a different corpus refer to, resulting in a clustering over these mentions.

---

[1] http://en.wikipedia.org/

Our approach to distant-labeling for coreference consists of three steps. Given the set of mention strings, the first step performs preliminary within-document coreference to resolve entities for mentions that appear in the same document. Second, a set of candidate entities for each mention string is identified from Wikipedia using the redirect and disambiguation pages. This set of candidates is reduced using a relevance scoring function based on a generative model over the article and Wikipedia pages. The steps are described in detail below in the context of person-name disambiguation in newswire articles, however the method can be applied to other forms of cross-doc coreference.

### 3.1  Within-Document Coreference

To reduce the number of mentions available for cross-document, within-document coreference is usually applied as a preprocessing step to resolve the entities within the document [1,6,29]. The task of resolving proper nouns within a document is usually straightforward; however, incorporating pronouns and nominal nouns (like "he", "she", "the president", etc.) make within-doc coreference considerably more challenging.

Since our method of within-doc coreference is applied only to mention strings that are proper nouns, it is similar to that used in [6,27]. A distance function is defined between a pair of mention sets (clusters) that uses hand-tuned weights over features that look at various string matches (such as whether the first name/last name is same) and gender matches (a mention contains "Mr." and the other contains "Mrs."). We apply standard greedy agglomerative clustering using this distance function, and use a low threshold to obtain high precision. Although this may result in multiple within-doc entities that refer to the same entity, we rely on the cross-doc disambiguation to resolve these. For the rest of the paper we refer to the within-doc coreference entities as mentions, and the strings as as sub-mentions. For each mention, the longest sub-mention is selected as its canonical string representation.

This within-doc coreference system is simple and domain-specific. The rest of the approach does not rely on a particular choice of within-doc coreference and instead a machine-learned coreference model that also considers pronouns can be used, such as [9]. However, since it is not directly relevant for cross-document coreference, for the purposes of illustration in this paper we used the domain-specific approach.

### 3.2  Identifying Candidate Entities

For each canonical mention string, a set of potential Wikipedia entities that the mention may refer to need to be identified. To discover these candidates, we utilize the *redirect* and *disambiguation* pages available in Wikipedia.

*Redirect* pages are used to forward the user to the page with the correct title, given that the user query is either a common misspelling ("Barak Obama"), an alternate spelling ("Dick Nixon"), or refers to multiple entities of which one is prominent ("Obama"). Redirect pages also exist for less common but difficult to disambiguate cases of renames (Sean Combs/Puff Daddy), typographical issues (Pointcare/Poincaré), spacing and casing (vangogh/Vincent van Gogh), etc. Note that the *destination* of a redirect link is not always a content page, it may be another redirect page. These redirections offer a reliable signal for identifying a candidate entity for a mention, since many different variations of the entity are represented.

Although the redirect pages provide alternate mentions for a single entity, often the same string can refer to multiple entities, for example, "Hillary" may refer to *Hillary Clinton* or to *Edmund Hillary*. To incorporate this information, Wikipedia includes *disambiguation* pages that list, for a given string, all the entities on Wikipedia that the string may refer to. These pages can range from 2 or 3 entities ("William Clinton") to more than 50 entities ("John Smith"). We use these disambiguation pages to expand the set of candidate entities for a given string.

The set of candidate Wikipedia pages for a given mention $m$ is computed as follows. We start by initializing a set $S = \{m\}$. Within a loop, we go through every element of $s \in S$, and do the following:

1. if there is a redirect page $s \rightarrow s'$, we remove $s$ and insert $s'$ into $S$.
2. if there is a disambiguation page for $s$ containing $S'$ as the set of possible pages, we remove $s$ and update $S \leftarrow S \cup S'$.
3. if $s$ is a content page in Wikipedia, do nothing
4. if $s$ is not present in Wikipedia, remove it from $S$.

The above process is repeated until $S$ stops changing, resulting in a final set of Wikipedia page candidates for $m$. This algorithm is analogous to taking a transitive closure of a graph where the mentions and content pages are the nodes, and the redirect and disambiguation pages represent the edges.

Note that this approach may result in many candidates that are trivially inapplicable to the mention string; however, the objective is to obtain a super-set of candidates. Furthermore, we may discover only one candidate for a mention on Wikipedia, but it may be the incorrect one (i.e. the mention's correct entity does not exist on Wikipedia). The next section describes how this set of candidates is pruned to a single (or no) match.

### 3.3 Selection Using Multinomials

Once the candidates for each mention are identified, the method selects which candidate the mention refers to, or whether it refers to none of the Wikipedia entities. To make this decision, we use the contents of the Wikipedia page of the candidate and the news article that the mention belongs to. A score is calculated between the article and the Wikipedia page, and the candidate with the highest score is picked (or none of the candidates are picked if this score is below a threshold). It is possible to use an alternative approach that only considers the local contexts around the sub-mentions instead of the complete article text, however we are interested in *thematic* match between the candidate page and the mention, and the local contexts are a noisy signal of this match.

The task of identifying the best candidate for a mention is represented as a retrieval problem. In information retrieval, documents are ranked for a given query according to the probability that the query is generated from the document. The document is represented as a unigram language model (multinomial over the words) and the probability of generating the query is the product of the probabilities of generating individual tokens of the query. Smoothing is often introduced as a back-off distribution so that tokens that do not appear in the document have a small positive probability of being generated [25,32].

Since the candidates refer to a single Wikipedia entity, and an article may refer to multiple entities, we rank the candidate pages (queries) according to the probability that they were generated from the news article (document). Tokens are weighted according to the standard *inverse-document frequency*[2] so that the rare tokens contribute more to the generation probability. To account for the back-off distribution, we use uniform multinomial for the global language model ($P(t|M_g) = \alpha$), and use linear interpolation (Jelinek-Mercer smoothing) using $\lambda$ ([17]).

The model is described below, where $c$ is the candidate page, $a$ is the article, $M_a$ is the language model for article $a$, $M_g$ is the global language model, $n_{t,i}$ is the count of word $t$ in page/article $i$, and $idf_D(t)$ is the inverse document frequency for word $t$ in corpus $D$.

$$P(c|a) \propto \prod_{t \in c} P(t|a)$$

$$P(t|a) = (\lambda P(t|M_a) + (1-\lambda) P(t|M_g))^{w_c^t}$$

$$= \left(\lambda w_a^t + (1-\lambda)\alpha\right)^{w_c^t}$$

$$\text{where } w_c^t = \frac{n_{t,c} \times idf_C(t)}{\sum_{t \in c} n_{t,c} \times idf_C(t)},$$

$$\text{and } w_a^t = \frac{n_{t,a} \times idf_A(t)}{\sum_{t \in a} n_{t,a} \times idf_A(t)}.$$

We rank the candidates according to $P(c|a)$ based on the article that contains the mention. To account for the case where the mention does not refer to any of the candidates, a threshold $\beta$ is used on the probability. Therefore we apply our candidate selection method even when mentions have a single candidate.

### 3.4 Evaluation

To evaluate the quality of the labels, we apply our method to the New York Times corpus [30] which contains 20 years of New York Times articles. Our approach is applied to a subset (Jan 1, 2004–June 19, 2007) that consists of 308k articles. The Stanford NER tagger [12] is used to extract person-name string from these articles, resulting in 5 million sub-mentions. High-precision within-doc coreference identifies 2.5 million mentions that are used in cross-document disambiguation. We use a recent snapshot of Wikipedia[3] consisting of 6.2 million pages, out of which 2.5 million are redirects and 121k are disambiguation pages.

---

[2] Since the distribution of the words in Wikipedia and our corpus may differ considerably, we compute separate *idf*s.
[3] enwiki-20080103-pages-articles.xml

Distantly Labeling Data for Large Scale Cross-Document Coreference        7**Table 1.** Distribution of Entity Sizes

| Entity Size | Number of Entities |
|---|---|
| 1 | 50,654 |
| 2-9 | 55,868 |
| 10-99 | 17,976 |
| 100-999 | 1,338 |
| 1000-9999 | 31 |

**Table 2.** Accuracy of the Labels

| Subset of Mentions | Correct Decisions / Total Mentions | Accuracy |
|---|---|---|
| Single Candidate | 872 / 888 | 98.2 |
| Multiple Candidates | 153 / 219 | 69.9 |
| Overall | 1025 / 1107 | **92.6** |

Out of the 2.5 million mentions, our method of candidate set selection does not find any candidates for $\sim 1.4$ million mentions, due to occurrence of rare entities, absence of sufficient context and string variations that are not captured by the *redirect* pages. Since we rely on Wikipedia for our distant-labels, these mentions cannot be resolved and are hence ignored.

Since we want as little smoothing as possible, we set $\alpha = 10^{-4}$ and $\lambda = 1 - \alpha$. This results in higher weights to rare word matches (and less weight to common word matches). The threshold for rejecting candidates $\beta$ is set to $e^{-18}$. Running our method of entity selection on $\sim 1.1$ million mentions (172k unique strings) results in matches to 125k entities (Wikipedia pages). We show the distribution of the resulting entity sizes in Table 1.

The creation of the dataset took a total of 23 hours, out of which 15 hours were for the NER tagging. Since there is no computation on pairs of mentions, our approach is linear in the size of the document corpus, and will scale to larger data, for e.g. all 20 years of the New York Times corpus. Furthermore, each decision is made independently, leading to parallelization across machines that can considerably reduce the running time.

We evaluate the accuracy of the resulting data set by manually examining the Wikipedia page selection decision made for individual mentions. $1,107$ total mentions are sampled randomly from the training data and their entity selection is labeled as correct or incorrect. The accuracy reported in Table 2 includes subsets of the labeled mentions that have only a single candidate (decision on whether to pick it or not) and ones that have multiple candidates.

The overall accuracy of $92.6\%$ for creating training data for coreference containing more than a million mentions without any manual intervention makes our approach useful for real-world applications. Since resolving disambiguation is a difficult task even by humans (some mentions had more than 50 candidates), we are impressed that our language model based method achieves an accuracy of $69.9\%$ for mentions with mul-tiple candidates. Furthermore, much of the dataset does not contain these disambiguations, resulting in the redirect link resolving a number of correct entities, achieving an accuracy of $98.2\%$ when there are no disambiguations.

### 3.5 Dataset Ambiguity

In this section we give some examples of cross-document coreference decisions that are difficult to achieve without our distant-labeling approach.

- **Renaming:** Consider the changing aliases of celebrities, for example mentions "Sean Combs" and "Puff Daddy". Since the contexts around these mentions vary significantly due to temporal differences, unsupervised methods that rely on the mention string or the contexts may not merge these mentions. Furthermore, these mentions are problematic for human labelers who lack the domain-specific knowledge required to disambiguate them. However, Wikipedia provides a redirect link, "Sean Combs→Puff Daddy". Similarly, for the example in Fig 1, "Lovebug Starski" appears on the disambiguation page for "Kevin Smith".
- **Alternate Names:** There are many examples of cases where the same first name is often referred to by alternate strings, such as "Richard/Dick", "Chuck/Charles", "Dan/Daniel", "Robert/Bob/Bobby", etc. Many approaches use a pre-built *gazetteers* for such renames, however they require domain expertise to build, may be incomplete, and may not always be applicable (for e.g. "Edward" is sometimes "Ted", as in Kennedy, but this is uncommon). Wikipedia redirect and disambiguation links provide alternatives only if the entity is sometimes referred to by the alternate name.
- **Same Names:** This is a major concern for cross-document coreference wherein same mention strings refer to different entities, e.g. "John Smith". Unsupervised techniques that rely on contexts can create more/less entities than truth. This problem is magnified when the different entities belong to similar fields ("Michael Moore", filmmaker, and "Michael D. Moore", film actor and director). From Wikipedia we know which entities to disambiguate between, and our relevance model uses the article text to select a candidate, resulting in an accuracy of $69.9\%$.
- **Singletons:** Many mention strings in news articles often refer to entities that do not appear in the corpus again (local opinion pieces, death notices, one-off criminal stories, etc.). Pruning these involves considerable labor, as not all articles in a certain category only refer to singleton entities. Frequency based analysis is also not enough, since many of these mentions share the same string with mentions from other articles. Our approach eliminates these from Wikipedia candidates based on the threshold on the article-page scoring function. However, as a side effect, this method does not resolve isolated references to prominent entities, e.g. reference to "George Bush" in a sports article.

We investigate the level of ambiguity in our generated dataset by analyzing precision and recall scores of pairwise coreference decisions for some simple baseline procedures evaluated using our generated labeled data as ground truth (see Table 3). Precision, in context of clustering, refers to the ratio of the number of pairs of mention that were correctly predicted to be coreferent (*true positives*) to the total pairs of mentions that were predicted to be coreferent (*true positives* + *false positives*). Precision thus encourages smaller clusters that are of high quality. Recall, on the other hand, refers to the number of true positives to the number of pairs of mentions that are coreferent as per the truth (*true positives* + *false negatives*), and encourages bigger clusters that are super-sets of true clusters. Precision and recall scores of some simple baseline clustering techniques (by treating our generated data as truth) gives us insight into the ambiguity in our data.

**Table 3.** Ambiguity Analysis using Pairwise Decisions

| Pairwise Metric (%) | Unique Name | Last Name |
|---|---|---|
| Precision | 97.29 | 35.49 |
| Recall | 81.49 | 98.85 |
| F1 score | 88.69 | 52.14 |

First we cluster the mentions according to their unique canonical strings, that is, each cluster contains mentions whose canonical names are string-identical. There are two important observations here. First, the precision is below 100% revealing that unique name clustering is merging names that refer to different entities, that is, demonstrating "same names" ambiguity. Second, the recall is considerably lower indicating that entities are being referred to by multiple strings, demonstrating "renaming" and "alternate names" ambiguity.

Next, we relax the notion of canonical name clustering to a *last name* clustering, where mentions are in the same cluster if their last names are string-identical. If there were no "rename" ambiguity, then we would expect perfect recall (100%) for the last name clustering. However, this is not the case, implying that there is a substantial amount of "rename" ambiguity in our dataset (although considerably less than the "alternate name" ambiguity).

This pairwise analysis, along with the manual evaluation in Table 2, can be used to estimate the quality and frequency of the various types of disambiguations in the dataset, and therefore its utility as a cross-document coreference corpus.

## 4 Cross-Document Coreference

Now that we have described our distantly labeled coreference corpus, we demonstrate its utility for training a sophisticated discriminative model of coreference. Since this conditional random field based model only consists of features based on the raw text surrounding the mentions, and does not use any supervision from Wikipedia, it can be applied to any source of text. In particular we reveal that the model is capable of generalizing to a held-out evaluation set that contains mentions and entities that are not encountered during training.

### 4.1 Model

We adapt the state-of-the-art within document coreference model of [9] to our problem setting. In particular, we use a conditional random field with features defined over mention-pairs and cross-document entities. There are two types of hidden variables: mentions (denoted $m_i$) and entities (denoted $e_i$). The mention variable has a domain that ranges over possible entities, and the entity is a set-valued variable whose domain ranges over all possible subsets of mentions. Further, there are variables that encapsulate all the observed properties of mentions, which are implicitly part of the hidden mention variable. These properties include the mention text, the canonical representation for the mention, and the bags of words extracted from context windows as well as the sub-mention texts. These factors model the dependencies between mention pairs in the same cluster ($\psi^+$ modeling attraction), and mention pairs in different clusters ($\psi^-$ modeling repulsion). Additionally, we have a factor over each entity variable ($\psi$) that can examine the cohesiveness of a coreference cluster. Figure 2 shows our model instantiated with three mentions over a two-entity hypothesis.

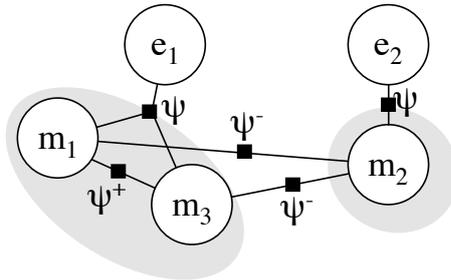

**Fig. 2.** The Factor graph for coreference with two entities and three mentions. The shaded areas represent the clustering.

Let $\mathcal{M}$ be the set of hidden mention variables, $\mathcal{E}$ be the set of hidden entity variables, $X$ be the set of observed mention properties, and $Y = \mathcal{M} \cup \mathcal{E}$ be the set of all hidden variables. Further let the notation $m_i.e$ denote the entity that $m_i$ references, let $P^+ = \{\langle m_i, m_j \rangle \in \mathcal{M}^2_{i<j} | m_i.e = m_j.e\}$ and similarly $P^- = \{\langle m_i, m_j \rangle \in \mathcal{M}^2_{i<j} | m_i.e \neq m_j.e\}$. Then, the probability distribution over the coreference hypothesis space (clusterings) is given as:

$$\pi(y|x) \propto \prod_{e \in \mathcal{E}} \psi(e) \times \prod_{P^+} \psi^+(m_i, m_j) \qquad (1)$$
$$\times \prod_{P^-} \psi^-(m_i, m_j)$$

where $\psi^\pm(m_i, m_j) = \exp\left(\theta^\pm \cdot \phi(m_i, m_j)\right)$ is a log-linear combination of parameters $\theta$ and the feature function $\phi$. Note that while the underlying feature-functions for these two types of factors are equivalent ($\phi$), the parameters between them are not tied, allowing separate sets of weights to be learned. Our model is implemented as an *imperatively-defined factor graph* using the FACTORIE probabilistic programming library [21].

### 4.2 Features

Since we want to apply the model to arbitrary sources of mentions, we rely only on the text, stored as observed properties of the mentions ($X$). As described above, these properties include the mention text, the canonical representation for the mention, and the bags of words extracted from context windows as well as the sub-mention texts.

Given a mention pair $\langle m_i, m_j \rangle$ we define the following binary pairwise features $\phi(m_i, m_j)$:

- **canonical match** true iff $m_i$'s canonical representation is lower-case string identical to $m_j$'s. Similarly, we include a **canonical mismatch** feature.
- **last name match** true iff $m_i$'s canonical last name is lower-case string identical to $m_j$'s Similarly, we include a **last name mismatch** feature.
- **cosine distance features** for each type of bag-of-words, the cosine distance is measured and quantized into ten bins.

For an entity $e$, we implement the following binary features

- **cluster size features:** thresholded cluster sizes ($== 1, > 1, > 2, > 4$)

### 4.3 Inference

In this section we describe a method for scaling inference to large datasets. In particular, we use a local search method based on the Metropolis-Hastings (MH) algorithm with a canopy-based proposer. We briefly describe MH and canopies, then present our jump function.

MH is a Markov-chain Monte Carlo method that stochastically performs local changes by probabilistically accepting jumps from a proposal distribution $\mathcal{Q}$ conditioned on the current state $y$ producing a new configuration $y'$. The proposed configuration $y'$ is accepted with probability $\alpha = \pi(y'|x)/\pi(y|x) \times q(y|y')/q(y'|y)$, where $\pi$ is the distribution encoded by the model (see Eq 1), and $q$ is the probability of proposing the jump to $y'$. Since we are performing *maximum a posteriori* (MAP) with no latent variables, we can safely ignore the ratio containing $q$ [21].

In order to avoid inefficiencies arising from unnecessary exploration, we inject the following domain-specific knowledge into the proposal distribution. First, the proposal distribution is designed so that inference explores only the space of valid configurations (settings to the hidden variables that result in an invalid clustering are not considered). Second, we use the idea of canopies to propose jumps that are more likely to be accepted by the model, and that lead to high-scoring configurations.

A canopy is a relaxation of a clustering where mentions can refer to more that one entity (in other words the transitivity assumption is not enforced and mentions can be in more than one cluster) [20]. Formally, we define a canopy $C$ as a *set of mention sets*. Typically, canopies are constructed so that mentions occurring in the same set are highly likely to be coreference, for example, they can be constructed such that all mentions in the same set are within a certain cosine threshold of each other.

Let $\Gamma = \{C_i\}$ be a set of canopies. Also, let the notation $t \sim_\rho T$ mean to draw an element $t$ from a set $T$ with probability distribution $\rho : T \to [0, 1]$ s.t. $\sum_{t \in T} \rho(s) = 1$. The class of proposal distributions based on canopies is defined as follows:

1: **Input:** current configuration $y$
2: **Output:** proposed configuration $y'$
3: $C \sim_\rho \Gamma$   //pick a random canopy
4: $S \sim_\rho C$   //pick a set of mentions from canopy
5: //pick a new entity for mention
   $m_a \sim_\rho S$
   $m_b \sim_\rho S$
   $m_a.e \leftarrow m_b.e$   // move $m_a$ to $m_b$'s entity
6: **return** $y'$

In practice we take $\rho$ to be uniform in distribution. In our implementation we use the *last name* and *canonical name* clusterings from Section 3.4 as the set of canopies $\Gamma$. We further enrich the above proposal distribution to create new entities and explore random configurations. Specifically, with 20% probability, we pick a random mention and move it to a random entity (this entity may be an empty one), and with 80% probability we run the canopy proposer.

Note that for each step, the jump function moves a single mention from one entity to another; each jump changes the settings of only three hidden variables. Therefore, even with factors over entire entities, evaluating the MH acceptance score requires computing as many factors as the number of total mentions in the two entities.

### 4.4 Learning

Parameter estimation is performed with SampleRank [34], an extremely efficient stochastic gradient-ascent based method that solves a ranking-objective function. SampleRank employs the same proposal distribution as used during inference and learns a model whose probabilities correspond with a user-specified ranking function over coreference configurations. In particular, we learn a model that ranks configurations according to pairwise accuracy.

### 4.5 Experiments

We present preliminary experiments demonstrating that the within document coreference model can be scaled to perform inference on a large dataset. In particular, we use our distantly-labeled NYT data consisting of over a million mentions, as described in Section 3.4. This is evenly divided into training and testing sets, each contains $\sim 550,000$ mentions and $\sim 90,000$ entities.

We perform ten iterations of SampleRank on the training set, where an iteration consists of $100,000$ Metropolis-Hastings steps; each iteration takes on average only $19.6$ minutes and training takes under five hours total. We perform inference using five million Metropolis-Hastings steps on the held out test data. Inference takes 9.5 hours and achieves a Pairwise F1 score of 89.83%. This high score is encouraging for a model that is trained on our distantly-labeled data. Note that since most of the entities that are present in the training data are absent from the evaluation, our model is resolving mentions to unseen entities. This provides evidence that our model may generalize to entities that are not present in Wikipedia, however we could not evaluate this since our evaluation data only consists of entities that appear in Wikipedia.

## 5  Related Work

Even though the cross-document coreference problem is challenging and lacks large labeled datasets, its ubiquitous role as a key component to many knowledge discovery tasks has inspired several efforts.

There are a number of unsupervised approaches to the problem, many of which rely on a scoring function for pairs of contexts that is used for clustering. One of the first approaches to cross-document coreference [1] uses a pre-trained within-document step, followed by an idf based scoring function for pairs of contexts for clustering. Ravin et al. [29] extend this work to be more scalable by comparing pairs of context only if the mentions are deemed ambiguous enough using a heuristic. Others have explored multiple methods of context similarity, and concluded that agglomerative clustering provides effective means of performing inference [14]. Pedersen et al. [24] and Purandare & Pedersen [28] integrate second-order co-occurrence of words into the similarity function. A number of other approaches include various forms of hand-tuned weights, dictionaries, and heuristics to define similarity for name disambiguation for clustering [6,2,27].

Since the unsupervised techniques make strong assumptions about the data and/or contain domain specific heuristics, techniques that rely on minimal supervision have been proposed. Mann & Yarowsky [18] extract biographical facts from the Web, such as birthdate, which are used as features for clustering. Niu et al. [23] incorporate information extraction into the context similarity model, and construct small annotated datasets to learn some of the parameters. There has been little work in completely supervised cross-document coreference. The only work of which we are aware is Finin et al. [11]. They incorporate features from Wikitology to train a pairwise classifier. Similarly, Mayfield et al. [19] use additional features that are information extraction based. Both these systems are trained on a small subset of the ACE 2008 data.

A number of techniques above also create datasets for evaluation. Bagga & Baldwin [1] generate a small, highly-ambiguous "john smith" dataset. Gooi & Allan [14] create the ambiguous "Person-X dataset" that replaces names of different entities to be the same *Person-X*. Although challenging and helpful for academic evaluation, these artificial datasets offer little realism. Niu et al.[23] generate approximate data sets for partial supervision, however the datasets are too small and noisy to be useful for training. A number of tools have been proposed to help annotators and the resulting datasets have been released. Bentivogli et al. [5] introduces a small dataset containing high ambiguity (209 names over 709 entities), but the data set is not big enough for large-scale models. Day et al.[10] also introduce a tool and a corpus, however the corpus offers little ambiguity.

Various aspects of Wikipedia have been used as supervision and features for a number of information extraction tasks. Features based on Wikipedia, including categorical and structure, have been used to train supervised models of within-document [26] and cross-document [11,19] coreference. Similar to our generative model, wikipedia has been used to score similarity between documents [13,16,33]. By making a "one person per document" assumption, Han & Zhao [16] treat document clustering as an approach to unsupervised coreference. Strube & Ponzetto [33] uses the semantic relatedness of articles as a feature for a supervised coreference model.

Our work is most similar to Bunescu & Pasca [7] and Cucerzan [8] which also use Wikipedia to disambiguate entities in an unsupervised manner. Bunescu & Pasca [7] use content and categorical information to create features for a scoring model that can disambiguate mentions that appear in wikipedia articles. This scoring model is trained on the existing links in Wikipedia. It unclear whether this method can generalize to mentions from other sources such as newswire. Cucerzan [8] disambiguates mentions in arbitrary data sources, and is evaluated on news articles. The vector representation of each document is created using features from wikipedia, and the dot product is used to denote document similarity. This disambiguation approach was applied to a small dataset. Additionally, none of these approaches train a cross-document coreference model that can run on a large number of mentions and resolve entities that do not appear in Wikipedia, restricting their utility.

## 6   Conclusions

Motivated by the difficulty of labeling data for large-scale cross-document coreference, we propose a distantly-labeling approach to automatically produce large datasets using Wikipedia. We applied the method to the New York Times corpus, and the noise and ambiguity in the generated dataset were analyzed. To enable cross-document coreference on this large dataset, a canopy-based sampling approach for training and inference was introduced. The model that we trained on this data has multiple uses in downstream applications, such as search, reputation analysis, trend analysis, etc. Furthermore, the predictions of the model can be used to suggest additional disambiguations and redirects for Wikipedia.

There are a number of avenues for future work. We intend to release the dataset so that the community of cross-document coreference can benefit from a large labeled corpus. Even though the current level of noise is acceptable, our method can be improved to create less noisy datasets, using more complicated models than the current. More ambiguity can be artificially introduced into the dataset; although this is not realistic, the resulting dataset may be more useful for evaluation of cross-document coreference methods.

## Acknowledgements

This work was supported in part by the Center for Intelligent Information Retrieval, in part by SRI International subcontract #27-001338 and ARFL prime contract #FA8750-09-C-0181, and in part by UPenn NSF medium IIS-0803847. Any opinions, findings and conclusions or recommendations expressed in this material are the authors' and do not necessarily reflect those of the sponsor.

Distantly Labeling Data for Large Scale Cross-Document Coreference 15